# Automatically Identifying Comparator Groups on Twitter for Digital Epidemiology of Pregnancy Outcomes

Ari Z. Klein, MA, PhD[1], Abeselom Gebreyesus[2], Graciela Gonzalez-Hernandez, MS, PhD[1]
[1]Department of Biostatistics, Epidemiology, and Informatics, Perelman School of Medicine, University of Pennsylvania, Philadelphia, PA, United States; [2]Department of Sociology, Anthropology, and Health Administration and Policy, University of Maryland, Baltimore County, Baltimore, Maryland, United States

**Abstract**

*Despite the prevalence of adverse pregnancy outcomes such as miscarriage, stillbirth, birth defects, and preterm birth, their causes are largely unknown. We seek to advance the use of social media for observational studies of pregnancy outcomes by developing a natural language processing pipeline for automatically identifying users from which to select comparator groups on Twitter. We annotated 2361 tweets by users who have announced their pregnancy on Twitter, which were used to train and evaluate supervised machine learning algorithms as a basis for automatically detecting women who have reported that their pregnancy had reached term and their baby was born at a normal weight. Upon further processing the tweet-level predictions of a majority voting-based ensemble classifier, the pipeline achieved a user-level $F_1$-score of 0.933, with a precision of 0.947 and a recall of 0.920. Our pipeline will be deployed to identify large comparator groups for studying pregnancy outcomes on Twitter.*

**Introduction**

Up to 22% of pregnancies result in fetal loss between 5 weeks and 20 weeks of gestation (i.e., miscarriage)[1], and nearly 1% of pregnancies result in fetal loss at 20 weeks or more of gestation (i.e., stillbirth)[2]. In the United States, 17% of pregnancies end in fetal loss[3]. Birth defects affect approximately 3% of live births[4] and are the leading cause of infant mortality in the United States[5]. Preterm birth (i.e., birth before 37 weeks of gestation) affects approximately 10% of live births in the United States[6] and is the leading cause of neonatal death globally[7]. Preterm birth with low birthweight (i.e., less than 5 pounds and 8 ounces) is the second leading cause of infant mortality in the United States[5]. Given that the causes of miscarriage[8], stillbirth[9], birth defects[10], and preterm birth[11] are largely unknown, additional sources of data for studying pregnancy outcomes should be explored in order to ultimately develop interventions.

Considering that 40% of adults between ages 18-29 in the United States use Twitter[12], in recent work, we identified a cohort of women reporting that their child has a birth defect[13], from a database containing more than 400 million publicly available tweets posted by more than 100,000 users who have announced their pregnancy on Twitter[14]. We used their timelines—the publicly available tweets posted by a user over time—to conduct an observational case-control study[15], in which we compared select risk factors among the women reporting a birth defect outcome (cases) and women for whom we did not detect a birth defect outcome, selected from the same database (controls). The study found that reports of medication exposure[16] during pregnancy were statistically significantly greater among the cases than the controls. More generally, the study demonstrates that social media data can be utilized as a complementary resource for epidemiology of pregnancy outcomes.

Because our database contains the timelines of women who have announced their pregnancy on Twitter, it provides a population from which to select internal comparator groups[17]. However, whereas comparator selection is traditionally facilitated by structured data (e.g., hospital and population-based registries[18]), social media requires identifying pregnancies for comparison from large amounts of noisy, unstructured text. Due to this challenge, our case-control study had selected women for comparison merely if we did not detect tweets in their timelines reporting a birth defect outcome. Thus, the comparator group might have included women with miscarriage, stillbirth, preterm birth, or low birthweight outcomes. Furthermore, the extensive amount of time needed to manually verify that the timelines of women reporting a pregnancy actually contained the timeframe in which they were pregnant prevented us from identifying more than one control per case.

In this paper, we present a natural language processing (NLP) pipeline for automatically identifying users from which to select comparator groups for observational studies of pregnancy outcomes on Twitter. Our approach is to automatically detect women who have reported that their pregnancy had reached term (i.e., at least 37 weeks of gestation) and their baby was born at a normal weight (i.e., at least 5 pounds and 8 ounces). Because this approach would systematically exclude women from the comparator group if they had miscarriage, stillbirth, preterm birth, or low birthweight outcomes, it will enhance the validity of observational studies of adverse pregnancy outcomes on Twitter. This approach will also facilitate the selection of larger comparator groups by identifying women whose reports of a term pregnancy (e.g., *I'm 38 weeks pregnant*) would usually entail the availability of tweets posted during pregnancy.

**Methods**

*Data Collection*

We handcrafted regular expressions—search patterns designed to automatically match combinations of text strings—to retrieve tweets from our database[14] that potentially mention a term pregnancy or normal birthweight. To enhance the retrieval of posts on social media, we automatically generated lexical variants[19] (e.g., misspellings, abbreviations) of the keywords in the regular expressions. We designed the regular expressions based on high-precision query patterns because, for automated comparator selection, it is important to identify women who have actually reported a term pregnancy and normal birthweight. Ignoring retweets (i.e., tweets beginning with "RT @"), the regular expressions (with the lexical variants) retrieved 16,548 tweets, posted by 11,305 users. Table 1 presents simplified forms of the regular expressions (i.e., without the lexical variants), the number of tweets retrieved by each query pattern, and samples of (slightly modified) matching tweets, which provide some examples of the lexical variants. Some of the tweets matched multiple query patterns. The first five query patterns in Table 1 are for detecting a term pregnancy, and the last two are for detecting a normal birthweight.

**Table 1.** Regular expressions to retrieve tweets about a term pregnancy (term) or normal birthweight (NB). The bold text indicates the string of the (slightly modified) sample tweet that matches the regular expression.

| | Query Pattern | Outcome | Tweets |
|---|---|---|---|
| 1 | (?<!until\s)(?<!like\s)(?<!when\s)\bi(\W?m\|\sam) ((((3[7-9])\|(4[0-2]))\W?weeks)\|(full\W?term))<br>• **I'm 39 wks** pregnant with a 2 year old and feel guilty for relaxing in the afternoon! | Term | 4239 |
| 2 | (?<!be\s)(?<!&\s)(?<!and\s)(?<!was\s)\b(((\d\|(1[0-9])\|(2[0-1]))\W?days)\|([1-3]\W?weeks)) (until\|(away from)\|from\|to) (my due date\|i(\W?m\|\sam) due)\b<br>• I have **9days til my due date** | Term | 3443 |
| 3 | \b(my due date(\W?s\|\sis)\|i(\W?m\|\sam) due) (in\s)?(((\d\|(1[0-9])\|(2[0-1]))\W?days)\|([1-3]\W?weeks))(?!(\sfrom\|\ssooner\|\sbefore\|\safter))<br>• **I am due in 3 days***sssss* | Term | 1522 |
| 4 | (?<!from\s)\b(((tomorrow\|today)(\W?s\|\sis))\|(yesterday was)) my due date<br>• Can't believe **2days my due date**! Getting induced Wednesday. Can't believe it's over | Term | 1395 |
| 5 | \b((my due date(\W?s\|\sis)\|i(\W?m\|\sam) due) (tomorrow\|today)\|(my due date was yesterday)\|(i was due yesterday))<br>• This is supposed to be my "last" dr. appt. since **I'm due tmrw** | Term | 1167 |
| 6 | \b(born\|birth\|delivered\|arrived\|came\|meet\|welcome\|is.*here\|introducing\|debut\|entrance)\b.* \b((5\W?pounds\.?(\s?(\W\|&\|and)\s?)?([8-9]\|(1[0-5]))\W?ounces)\|(([6-9]\|10)\W?pounds\.?(\s?(\W\|&\|and)\s?)?([0-9]\|(1[0-5]))\W?ounces))<br>• **Meet our beautiful son. 8 pounds and 10 ounces**! We are so blessed | NB | 4764 |
| 7 | \b(born)\b.* \b((((2\.[5-9])\|([3-4]\.[0-9]))\W?kilograms)\|((2,?[5-9][0-9][0-9])\|([3-4],?[0-9][0-9][0-9]))\W?grams)\b<br>• Excited & proud to introduce our daughter. **Born on 20/01/16 at 2:43pm. 3.8kg**, 52cm long | NB | 41 |

Among the 16,548 tweets retrieved by the regular expressions, 2683 tweets were posted by 853 users who posted at least one tweet matching a "term pregnancy" pattern and at least one tweet matching a "normal birthweight" pattern. We expected that women reporting a term pregnancy and normal birthweight for their own pregnancy would post only a small number of matching tweets, so we generated frequency distributions of the tweets posted by each of the 853 users in order to identify bots, organizations, forums, or other types of user accounts that are not reporting personal information. Table 2 presents examples of typical tweets by four users who posted significantly more tweets than the average number of tweets posted by the 853 users. Using rules to remove all tweets by users who have posted a tweet beginning with *ccb*, *baby club update*, *from our inbox*, *a question from our inbox*, *fq*, *fan share*, *#fanquestion*, or *mummy to be advice*, six users (322 tweets) were removed. These linguistic patterns identify users that are not posting information about their own pregnancy. In the pipeline, pre-filtering rules will be used to exclude such users from the comparator group because many of their tweets would pose challenges for automatic classification.

**Table 2.** Sample tweets by users who posted significantly more tweets than the average number of tweets by users who posted at least one tweet matching a "term pregnancy" query pattern and at least one tweet matching a "normal birthweight" query pattern.

| User | Tweet | Frequency |
|---|---|---|
| A | CCB is thrilled to welcome another 8 lb., 11 oz. bundle of love into the world. Mother and daughter are bringing in 2013 with big smiles! | 121 |
| A | Baby Club Update: CCB is thrilled to welcome another 7 lb., 4 oz, bundle of joy into the world. Both moms and their son are doing great! | 121 |
| B | From our inbox: "I'm 40 weeks and 5 days with my third child, and my OB wants to induce me at 41 weeks 1 day. I... [URL] | 119 |
| B | A question from our inbox: "Are there any risks to using Evening Primrose Oil when trying for VBAC? I'm 39 weeks... [URL] | 119 |
| C | Fq: Please post anon. I'm 38 weeks 1 day today. I went to the doctor last week I was not dilated at all. Today I... [URL] | 54 |
| C | Fan share: Happy to let everyone know that I have given birth to a beautiful 6 lb, 4oz and 19 1/2 in long little... http://t.co/c3ujpLEL71 | 54 |
| D | #FanQuestion ~ I'm 37 weeks pregnant with my first baby and I'm having a boy. Could anyone suggest some nice but unusual names? Thanks x | 13 |
| D | Mummy to be advice ~ Today is my due date & no sign of little one making. How late were you & what did you do to speed the process up? | 13 |

*Annotation*

After pre-filtering the 2683 tweets retrieved by the regular expressions, two professional annotators manually annotated 2361 tweets by 847 users who posted at least one tweet matching a "term pregnancy" query pattern and at least one tweet matching a "normal birthweight" query pattern. Annotation guidelines were developed to help them distinguish two classes of tweets:

- *Positive*: The tweet indicates that the user's pregnancy had reached term or that the user's baby was born at a normal weight.

- *Negative*: The tweet does not indicate that the user's pregnancy had reached term or that the user's baby was born at a normal weight. Alternatively, the tweet indicates an adverse pregnancy outcome, despite a term pregnancy or normal birthweight.

To account for the possibility that a user's timeline spans multiple pregnancies, for women who have posted at least one "term pregnancy" tweet that was annotated as "positive" and at least one "normal birthweight" tweet that was annotated as "positive," two annotators used the posting dates of their "positive" tweets to help identify whether or not the women are reporting a term pregnancy and normal birthweight for the *same* pregnancy.

*Classification*

We used the 2361 annotated tweets in experiments to train and evaluate supervised machine learning algorithms. For the classifiers, we used the default implementations of (1) ZeroR, (2) J48 Decision Tree (J48), and (3) Logistic Regression (LR) in Weka 3.8.2, (4) the WLSVM Weka integration of the LibSVM implementation of Support Vector Machine (SVM), and (5) a majority voting-based ensemble of J48, LR, and SVM. ZeroR is a rule-based, baseline classifier that predicts every instance (i.e., tweet) as the majority class. For training, we used 1851 tweets posted by a random sample of 80% (677) of the 847 users. For evaluation, we used 510 tweets posted by 20% (170) of the users as a held-out test set. We stratified the sets based on the distribution of (1) users reporting a term pregnancy and normal birthweight for the same pregnancy, (2) users reporting a term pregnancy and normal birthweight for different pregnancies, and (3) users who did not post at least one "term pregnancy" tweet that was annotated as "positive" and at least one "normal birthweight" tweet that was annotated as "positive."

We performed text pre-processing prior to automatic classification. First, we normalized user names (i.e., strings beginning with "@") and URLs in the tweets. Then, we lowercased the tweets and normalized the lexical and semantic variants of a term pregnancy (e.g., *38 weeks*, *40 wks*, *full term*) and normal birthweight (e.g., *7 lbs 8 oz*, *6lbs & 15ozs*,

*5 pounds, 12 ounces*, *2.8kg*, *4,444 grams*) matched by the regular expressions, textually representing them as "_term_" and "_normalbirthweight_", respectively. As a preliminary approach to normalizing the names of babies mentioned in the tweets, we compiled a lexicon to detect the 400 most popular boys and girls names in the United States between 2010 and the present, identified by the Social Security Administration[20]. Finally, we removed non-alphabetical characters and stemmed[21] the tweets. Following pre-processing, we converted the tweets to word vectors and used Weka's NGram Tokenizer to extract word n-grams (n = 1, 2, 3) as features. We also used word clusters[22] as features, which provide generalized semantic representations of the words in the raw tweets; words appearing in similar collocate contexts (e.g., misspellings) are represented by the same cluster number.

For the ZeroR, J48, and LR classifiers, we used the default parameters in Weka. For the SVM classifier, we used the radial basis function (RBF) kernel and, based on performing 10-fold cross validation over the training set, set the *cost* parameter at $c = 128$ and the class weights at $w = 1.0$ for the "positive" class and $w = 8.5$ for the "negative" class. In order to automatically identify women reporting a term pregnancy and normal birthweight for the same pregnancy, we used the classifiers' predictions from 10-fold cross validation to identify an optimal temporal proximity between a user's "term pregnancy" and "normal birthweight" tweets that were predicted as "positive." We used the *python-dateutil* package to compute the number of days between the posting dates of a user's "term pregnancy" tweets that were predicted as "positive" and her "normal birthweight" tweets that were predicted as "positive." We set the threshold to 50 days, meaning that a user would be included in the comparator group if the classifier predicts a "term pregnancy" tweet as "positive" and a "normal birthweight" tweet as "positive" that the user had posted within 50 days of each other. Figure 1 illustrates our automatic pipeline.

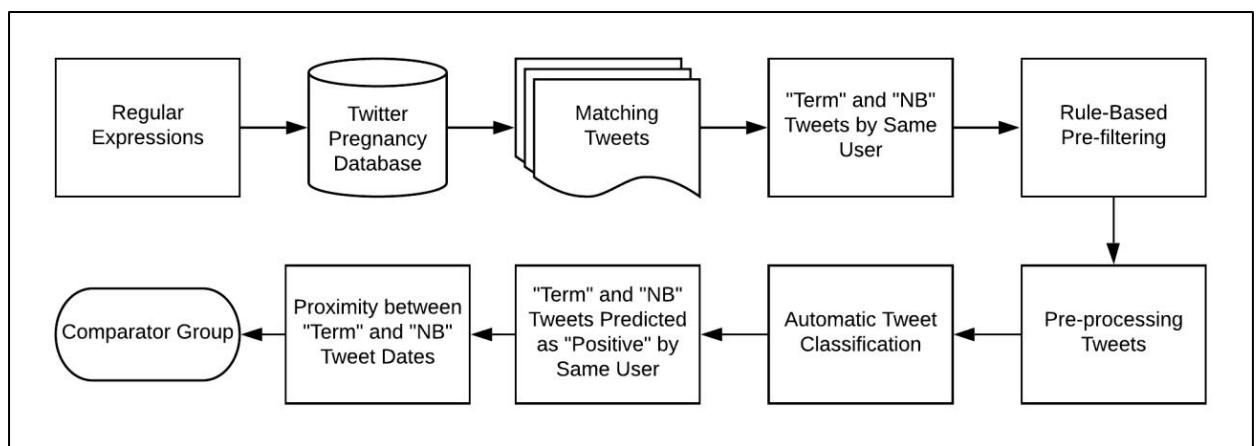

**Figure 1.** Automatic NLP pipeline for detecting women who have reported in tweets that their pregnancy had reached term (term) and their baby was born at a normal birthweight (NB), from which to select comparator groups for observational studies of pregnancy outcomes on Twitter.

### Results

*Annotation*

Two professional annotators manually annotated 2361 tweets by 847 users who posted at least one tweet matching a "term pregnancy" query pattern and at least one tweet matching a "normal birthweight" query pattern. Their inter-annotator agreement, based on 2170 overlapping annotations, was $\kappa = 0.80$ (Cohen's kappa), considered "substantial agreement"[23]. Upon resolving the disagreements, 2127 (90.1%) tweets were annotated as "positive" and 234 (9.9%) as "negative." Table 3 provides (slightly modified) examples of "positive" and "negative" tweets. Among these 847 users, 749 (88.4%) posted at least one "term pregnancy" tweet that was annotated as "positive" and at least one "normal birthweight" tweet that was annotated as "positive". For these 749 users, two annotators used the posting dates of their "positive" tweets to determine whether or not they refer to the *same* pregnancy, agreeing on 707 (94.4%) of the users. Upon resolving the disagreements, 683 (80.6%) of the 847 users were identified as reporting a term pregnancy and normal birthweight for the same pregnancy: "positive" users.

**Table 3.** Sample (slightly modified) tweets manually annotated as "positive" (+) or "negative" (-), with a "term pregnancy" (term) tweet and "normal birthweight" (NB) tweet per user.

| User | Tweet | Outcome | Date | Tweet +/- | User +/- |
|---|---|---|---|---|---|
| A | 17 days till my due date, so sad its going to be over but cant wait to see my baby's beautiful face😊💗 | Term | 2015-09-21 | + | - |
| A | last month i was in the most horrible pain delivering my baby. 1 month later i have a happy & healthy 9lb 7oz baby😊💗 | NB | 2015-10-28 | - | |
| B | I'm 38 weeks pregnant! Wow! I've never been this pregnant 😩😄 | Term | 2016-10-11 | + | - |
| B | Today is #WorldPrematurityDay. My 36 weeker was born weighing 6lbs5ozs and is now a healthy 18 month old! | NB | 2015-11-17 | - | |
| C | Today's my due date, but life clearly had other plans #5weeksold today! | Term | 2016-04-01 | - | - |
| C | My sister-in-law had her baby! Born January 22 at 8:44pm - 8lbs. 4oz. We really appreciate our first year anniversary gift! | NB | 2010-01-23 | - | |
| D | mine still fits over my 39+ week belly with some room to spare. (today is my due date). | Term | 2016-05-31 | + | - |
| D | My baby was born on Saturday 8/2 @ 2:45pm. 7lbs. 3oz. 20 in. | NB | 2014-08-05 | + | |
| E | I'm 2 weeks away from my due date. I've gotta keep my blood pressure normal 😁 | Term | 2016-09-17 | + | + |
| E | Meet my darling. Born at 17:15, weighing 2.8kg. Mom is in loads of pain | NB | 2016-10-12 | + | |

Table 3 provides pairs of "term pregnancy" tweets and "normal birthweight" tweets posted by the same users. The "term pregnancy" tweet by A was annotated as "positive" because *17 days till my due date* indicates that her pregnancy had reached at least 37 weeks of gestation. Although her "normal birthweight" tweet mentions a weight (*9lb 7oz*) that would be normal if this were her baby's weight at birth, the tweet indicates that this is how much her baby weighs at one month old, so it was annotated as "negative." Thus, A should be excluded from the comparator group. Although B does report that her baby was born at a normal weight (*was born at 6lbs5ozs*), the tweet was annotated as "negative" because it also indicates that the baby was born preterm (*my 36 weeker*). Thus, B should also be excluded from the comparator group. Even if B's "normal birthweight" tweet did not report a preterm birth and were annotated as "positive," B should still be excluded because the posting dates indicate that her tweets do not refer to the same pregnancy.

The "term pregnancy" tweet by C, at first, seems to report that her pregnancy had reached full term (*Today's my due date*), but it goes on to indicate that her baby had already been born and, moreover, is already 5 weeks old, so it was annotated as "negative" because her baby actually was born preterm. The "normal birthweight" tweet by C was annotated as "negative" because it refers to the birthweight of someone else's baby. The "term pregnancy" and "normal birthweight" tweets by D were both annotated as "positive," but the posting dates indicate that the tweets refer to different pregnancies, so D should be excluded from the comparator group. Finally, E should be included for comparator selection because her "term pregnancy" tweet indicates that her pregnancy had reached term (*I'm 2 weeks away from my due date*), her "normal birthweight" tweet indicates that her baby was born weighing at least 5 pounds and 8 ounces (*my darling…born…weighing 2.8kg*), and the posting dates indicate that her tweets refer to the same pregnancy.

*Classification*

Table 3 presents the precision, recall, and $F_1$-scores for automatic classification of "positive" tweets and users in the held-out test set, which contains 510 tweets posted by 170 users: 449 (88%) "positive" tweets and 61 (12%) "negative" tweets, posted by 137 users reporting a term pregnancy and normal birthweight for the same pregnancy, 13 users reporting a term pregnancy and normal birthweight for different pregnancies, and 80 users did not post at least one

"term pregnancy" tweet that was annotated as "positive" and at least one "normal birthweight" tweet that was annotated as "positive":

$$F_1\text{-score} = \frac{2 \times recall \times precision}{recall + precision}; \quad recall = \frac{true\ positives}{true\ positives + false\ negatives}; \quad precision = \frac{true\ positives}{true\ positives + false\ positives}$$

User-level performance is presented before and after applying the 50-day threshold between a user's "term pregnancy" and "normal birthweight" tweets that the classifier predicted as "positive." The user-level performance prior to applying the 50-day threshold is based only on the automatic classification of tweets. For all the classifiers, the 50-day threshold improved user-level performance. The ensemble classifier achieved the best $F_1$-scores at the tweet level (0.950) and the user level prior to applying the 50-day threshold (0.910). The ZeroR and ensemble classifiers achieved nearly identical $F_1$-scores at the user level after applying the 50-day threshold. Although the ZeroR classifier (0.934) performed marginally better than the ensemble classifier (0.933) overall, its precision (0.941) was slightly lower than the ensemble classifier's (0.947).

**Table 4.** Precision (P), recall (R), and $F_1$-scores (F) for automatic classification of "positive" tweets and users in a held-out test set of 510 tweets posted by 170 users.

| Classifier | Tweet P | Tweet R | Tweet F | User P[1] | User R[1] | User F[1] | User P[2] | User R[2] | User F[2] |
|---|---|---|---|---|---|---|---|---|---|
| ZeroR | 0.880 | **1.000** | 0.936 | 0.806 | **1.000** | 0.893 | 0.941 | **0.927** | **0.934** |
| J48 | 0.903 | 0.980 | 0.940 | 0.826 | 0.971 | 0.893 | 0.947 | 0.905 | 0.925 |
| LR | **0.931** | 0.935 | 0.933 | **0.860** | 0.898 | 0.879 | **0.958** | 0.832 | 0.891 |
| SVM | 0.896 | 0.996 | 0.943 | 0.820 | 0.993 | 0.898 | 0.940 | 0.920 | 0.930 |
| Ensemble | 0.910 | 0.993 | **0.950** | 0.840 | 0.993 | **0.910** | 0.947 | 0.920 | 0.933 |

[1] before applying the 50-day threshold to the classifier's tweet predictions
[2] after applying the 50-day threshold to the classifier's "positive" tweet predictions

**Discussion**

The $F_1$-scores for the best-performing classifiers—ZeroR (0.934) and a majority voting-based ensemble of J48, LR, and SVM classifiers (0.933)—represent a promising benchmark for automatically detecting women who have reported on Twitter that their pregnancy has reached term and their baby was born at a normal weight. As the confusion matrices in Figure 2 illustrate, the reason that the ZeroR classifier slightly outperformed the ensemble classifier after applying the 50-day threshold is twofold: after applying the threshold, (1) the ZeroR classifier successfully excluded all but one of the seven additional false positive users that had been misclassified based on its tweet predictions, and (2) the ZeroR classifier included one additional true positive user that was a false negative based on the ensemble classifier's tweet predictions. Still, we will pursue the ensemble classifier for comparator selection in practice (i.e., for identifying additional users in our constantly growing database[14]) because, as we will discuss, its user-level $F_1$-score will likely improve with additional pre-filtering rules, restrictions to the regular expressions, and annotated training data.

| Classifier | Before 50-Day Threshold | After 50-Day Threshold |
|---|---|---|
| ZeroR | Predicted<br>+   -<br>137   0   +   Actual<br>33   0   - | Predicted<br>+   -<br>127   10   +   Actual<br>8   25   - |
| Ensemble | Predicted<br>+   -<br>136   1   +   Actual<br>26   7   - | Predicted<br>+   -<br>126   11   +   Actual<br>7   26   - |

**Figure 2.** Confusion matrices for the user-level performance of the ZeroR and majority voting-based ensemble classifiers, before and after applying the 50-day threshold between a user's "term pregnancy" and "normal birthweight" tweets that the classifier predicted as "positive."

Error analysis of the ensemble classifier reveals that six of the seven false positive users stem from false positive tweets, and that three of the six users who posted false positives were accounts re-posting other users' tweets, such as the "term pregnancy" and "normal birthweight" tweets by A in Table 5. We will explore additional pre-filtering rules to remove tweets by such users prior to automatic classification. Many of the other false positive "term pregnancy" tweets report a future point at which the user's pregnancy will reach term, such as the "term pregnancy" tweet by B. Some of these errors can be addressed in the data collection module by incorporating additional "negative lookbehinds" into the regular expressions. Many of the false positive "normal birthweight" tweets report the birthweight of someone other than the user's baby, such as the "normal birthweight" tweet by C. Because these users' false positive tweets resulted in "term pregnancy" and "normal birthweight" tweets that were posted within 50 days of each other, such users would be incorrectly included in the comparator group.

**Table 5.** Samples of (slightly modified) tweets posted by false positive and false negative users.

| User | Tweet | Outcome | Date | Act. Tweet | Pred. Tweet | Act. User | Pred. User |
|---|---|---|---|---|---|---|---|
| A | Question from [name]: 'My question: I'm 37 weeks 3 days and planning to do my first home water birth... | TP | 2014-05-30 | - | + | - | + |
| | Massive congratulations to [name]!! [name] was born on June 1st at home weighing 7lb 4oz! | NB | 2014-06-03 | - | + | | |
| B | I still have a few weeks before I'm full term, but Idk if I'm gonna make it! Lol | TP | 2014-07-27 | - | + | - | + |
| | [name] was born 9 lbs 6 ounces. I love him so much already! He's perfect! #proudmommy | NB | 2014-08-30 | + | + | | |
| C | I'm 38 weeks pregnant today. | TP | 2017-09-06 | + | + | - | + |
| | [name] weighed 10 lbs 6 oz when he was born, and I weighed 9 lbs 2 oz. 😊 #iwantachubbybaby | NB | 2017-08-31 | - | + | | |
| D | Haven't done maternity pictures or a belly cast & I'll be 39 weeks tomorrow. | TP | 2015-09-26 | + | + | + | - |
| | At this time last year, my baby came into the world! 🎂🍼 [name]  10/6/2015 5:08 PM 7lbs 8oz | NB | 2016-10-06 | + | + | | |

Whereas the ensemble classifier misclassified tweets by nearly all of the false positive users, it correctly classified the tweets posted by nine of the eleven false negative users; in other words, most of the false negative users posted true positive "term pregnancy" and "normal birthweight" tweets that were more than 50 days apart, such as the "term pregnancy" and "normal birthweight" tweets by D in Table 5. While the 50-day threshold improved the user-level precision (and $F_1$-score) of the ensemble classifier, users such as D would be incorrectly excluded from the comparator group with the decline in recall—a trade-off we consider acceptable, given the importance of precision for automated comparator selection. However, in evaluating a higher, 125-day threshold directly on the annotated test set, we found that recall increased from 0.920 to 0.956, with an increase in precision (from 0.947 to 0.978) and $F_1$-score (from 0.933 to 0.967) as well. Thus, improving the performance of tweet-level classification and increasing the temporal threshold will improve user-level performance. As Figure 3 illustrates, the $F_1$-score for tweet-level classification may further improve with additional annotated training data.

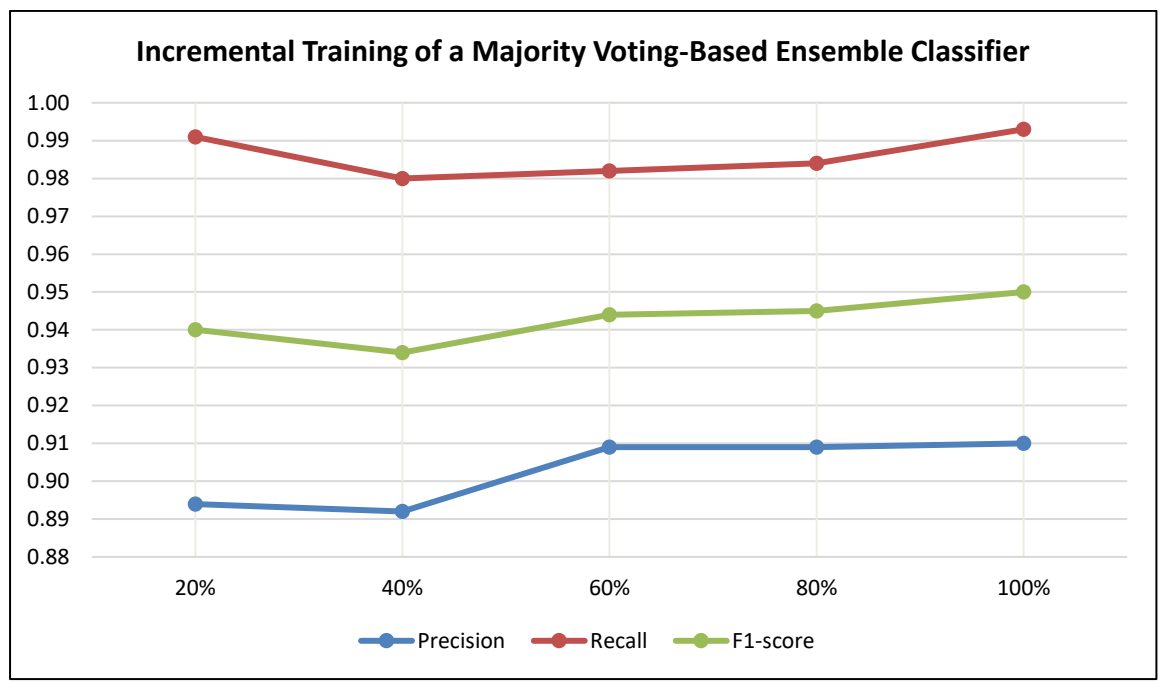

**Figure 3.** Precision, recall, and $F_1$-scores for the majority voting-based ensemble classifier, with incremental training sizes.

**Conclusion**

In this paper, we presented an NLP pipeline for automatically detecting women who have reported on Twitter that their pregnancy had reached term and their baby was born at a normal weight. Given the performance of the pipeline on a held-out test set (an $F_1$-score of 0.933), it will be deployed to identify large comparator groups for observational studies of birth defects, miscarriage, stillbirth, and preterm birth—the causes of which are largely unknown—on Twitter, advancing the use of social media as a complementary resource for studying pregnancy outcomes. Our analysis suggests that the performance of the pipeline may improve with additional pre-filtering rules, restrictions to the regular expressions, and annotated training data.

**Acknowledgements**

This work was funded by the National Institutes of Health (NIH) National Library of Medicine (NLM) grant number R01LM011176. The content is solely the responsibility of the authors and does not necessarily represent the views of the NIH or NLM. This study received an exempt determination by the Institutional Review Board (IRB) of the University of Pennsylvania, as it does not meet the definition of "human subject" according to 45 CRF § 46.102(f). The authors would like to acknowledge Karen O'Connor and Alexis Upshur for their annotation efforts, and Haitao Cai for performing database queries and developing code for modules of the pipeline.